\documentclass{article}
\usepackage{iclr2027_conference,times}

\usepackage{amsmath,amsfonts,bm}

\def\eqref#1{equation~\ref{#1}}

\def\1{\bm{1}}

\DeclareMathAlphabet{\mathsfit}{\encodingdefault}{\sfdefault}{m}{sl}
\SetMathAlphabet{\mathsfit}{bold}{\encodingdefault}{\sfdefault}{bx}{n}

\usepackage{algorithm}
\usepackage{algpseudocode}
\usepackage{amssymb}
\usepackage[disable]{todonotes}
\usepackage{hyperref}
\usepackage{url}
\usepackage{float}
\usepackage{xspace}

\newif\ifarxivversion
\arxivversiontrue 

\title{
Not All Thinking is Created Equal:
\\Latent Reasoning Discovers a Recurrent Search Algorithm for Depth Generalization
}
\author{Huzi Cheng\\
University of Minnesota\\
\texttt{hzcheng15@gmail.com} \\
\And
Zhewei Zhang\\
Independent Researcher \\
\texttt{zhzhewei36@gmail.com}
}
\newcommand{\btnlt}{\textbf{\emph{Bottleneck-latent}}\xspace}
\newcommand{\flt}{\textbf{\emph{Full-latent}}\xspace}
\newcommand{\drt}{\textbf{\emph{Direct}}\xspace}
\newcommand{\pstk}{\textbf{\emph{Pause-token}}\xspace}
\newcommand{\cotv}{\textbf{\emph{Chain-of-Thought}}\xspace}
\newcommand{\pqt}{ProsQA-Ext\xspace}

\ifarxivversion
\iclrfinalcopy
\fi
\begin{document}
\maketitle
\ifarxivversion
\lhead{}
\renewcommand{\headrulewidth}{0pt}
\fi

\begin{abstract}
Large Language Models can perform multi-step reasoning and improve task performance through different forms of intermediate computation, from token-based traces to computation carried out in latent space.
However, a question remains open: do these different forms of thinking rely on the same underlying mechanism?
To address this, we train and compare five variants of the same GPTNeoX backbone from scratch on an extended multi-hop reasoning task (ProsQA-Ext): a vanilla model, a Chain-of-Thought (CoT) model, a Pause Token model, and two latent-reasoning models that are optimized end-to-end without intermediate reasoning traces.
We find that, strong in-distribution (ID) performance does not guarantee depth generalization.
Vanilla, CoT, and Pause Token models solve ID problems well, but rely largely on local graph features and generalize poorly to out-of-distribution (OOD) problems with longer hops.
In contrast, latent variants generalize better and show internal dynamics consistent with forward reachability propagation on the graph.
Causal interventions and circuit analysis localize this computation to a sparse recurrent search circuit in the bottleneck latent model: an attention head retrieves graph relations, an MLP and the residual stream update the reachability state across recurrent steps, while multiple attention heads together then do the candidate matching.
Together, these results show that different thinking mechanisms can learn distinct computational solutions, even at similar ID performance.
In this setting, latent recurrence supports a reusable forward-search algorithm that generalizes beyond the training depth.

\end{abstract}

\section{Introduction}
Prepending  ``Let's think step by step'' to a prompt can improve pretrained language models' performance on reasoning tasks \citep{kojimaLargeLanguageModels2022, weiChainofThoughtPromptingElicits2022}.
Models that fail to answer directly can sometimes solve the same question by first generating intermediate steps.
More recently, this approach has been successfully scaled by baking the reasoning traces into the training rather than prompting
\citep{chungScalingInstructionFinetuned2024,hoLargeLanguageModelsReasoning2023,magisterTeachingSmallLanguage2023}, which enables smaller models to solve problems where CoT prompting alone is ineffective.

However, whether a model actually follows the reasoning traces it generates remains debated.
Part of the reasoning traces can be replaced or removed without hurting the final answer \citep{lanhamMeasuringFaithfulnessChainofThought2023,zhaoCanAhaMoments2026}.
Models can also benefit from intermediate steps using meaningless filler or pause tokens \citep{pfauLetsThinkDot2024,goyalThinkYouSpeak2024}.
Together, these findings suggest that useful intermediate computation need not be realized as a verbally meaningful reasoning trace.
Coconut \citep{haoTrainingLargeLanguage2025} and related works have demonstrated alternative ways to perform such computation by feeding high-dimensional vectors, instead of tokens, directly into the model \citep{weiSIMCoTSupervisedImplicit2025}. The mechanisms underlying this latent computation remain poorly understood.

Symbolic reasoning tasks were widely used to probe the circuits and computation inside language models \citep{wuHowDoTransformersLearn2025,brinkmannMechanisticAnalysisTransformer2024}.
\citet{zhuReasoningSuperpositionTheoretical2025} showed that latent thoughts theoretically can encode multiple search frontiers in superposition and enable parallel search. However, recent work finds that similar patterns also arise in models without recurrence and do not always causally affect the answer \citep{aswalObservablePatternsNotExplanations2026,rizvi-martelIllusionSuperpositionPrincipled2026}, leaving its causal role contested.
More broadly, it remains unclear how the learned computation differs across thinking interfaces, and what mechanisms support generalization beyond the training distribution.

To investigate this question, we focus on five model variants: a vanilla model, a CoT model, a Pause Token model, and two latent-reasoning models based on Coconut. One full-latent model retains access to all previous tokens, while the other bottleneck-latent model can only rely on the intermediate hidden representations when generating answers. We train these five models on an extended version of the well-established ProsQA task.
Notably, the latent variants are trained without intermediate reasoning traces and RL, allowing us to examine whether latent reasoning can discover a generalizable reasoning mechanism without being shown how to solve the task step by step and without slow trial-and-error process.

By testing these models on out-of-distribution (OOD) problems, we find that strong performance within the training range does not guarantee depth generalization, with the latent variants performing best on OOD problems. The strong generalization, together with the lack of shortcut effects in the latent models, indicates that they learn to reason rather than use surface heuristics.
Further causal interventions in the bottleneck-latent model show that intermediate states carry intermediate variables during forward search that are reused and transformed across recurrent steps.
We localize this computation to a sparse search circuit in which an attention head retrieves graph relations and an MLP, together with the residual stream, update the state for subsequent steps, and finally multiple attention heads read this information for candidate matching.
These findings show that different forms of thinking can learn very different computational solutions, even at similar performance.
Importantly, latent recurrence supports better discovery of a reusable computation that generalizes beyond the training depth.

\section{Methods}
\subsection{Task}
A signature of a model that understands rules and can reason with them is that it learns from small scale datasets and generalizes to unseen, more complex problems.
In natural language problems, multi-hop symbolic reasoning, such as extended syllogisms, is a good candidate for such datasets:
the level of difficulty, i.e., the number of hops, can be controlled, and the symbols used can be permuted without changing the meaning.
In this study, adapted from ProsQA by \citet{haoTrainingLargeLanguage2025}, we construct such a task, \pqt.
As shown in Fig.~\ref{fig:task-model-perf}A, each \pqt sample $(x,y) \in \mathcal{D}$ describes a directed acyclic graph (DAG) $G=(V,E)$ and a question about $G$.
The graph description $g$ is a token sequence of premises of the form \texttt{A is B.}, each representing a directed edge from A to B.
The query $q$ gives a root node $r\in V$ and two candidate nodes $c_0,c_1 \in V$.
Together, they form the complete input $x=g\Vert q$.
Exactly one candidate is reachable from $r$, at a shortest distance of $H$; the other is either an isolated node or lies on a chain whose root is not $r$.
We denote the reachable candidate by $c^*$ and the reference answer-token sequence by $y$, which states that $r$ is $c^*$.
At a fixed $H$, node labels, premise order, and candidate positions are randomly sampled, yielding varied graph structures and inputs $x$, while the underlying reachability operation $f$ remains unchanged.
Unlike ProsQA, we carefully control the generation of $G$ so that no superficial features can be exploited to infer $c^*$ ($\approx 50\%$ accuracy), and we use two separate datasets: a training set with $H \in \{3,\ldots,6\}$ and a validation set with $H \in \{7,\ldots,12\}$.

\subsection{Models}
\begin{figure}[htbp]
\centering
\includegraphics[width=0.3\linewidth]{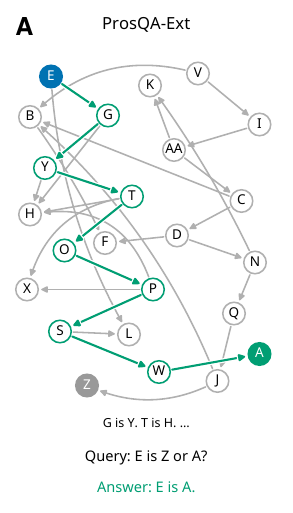}%
\includegraphics[width=0.7\linewidth]{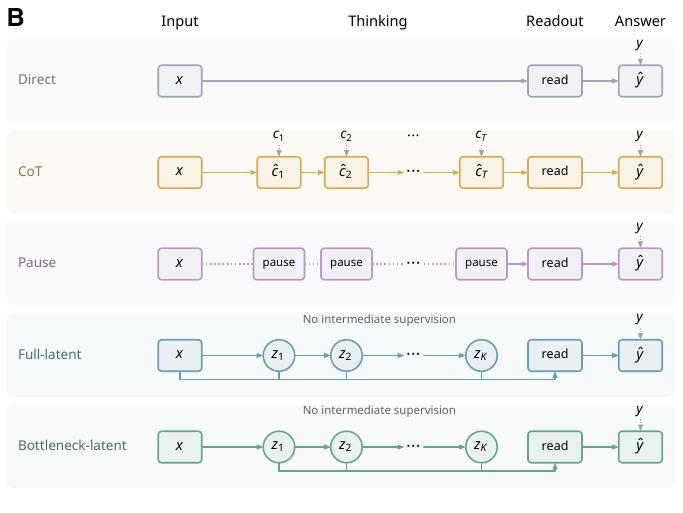}
\vspace{0.5em}
\includegraphics[width=\linewidth]{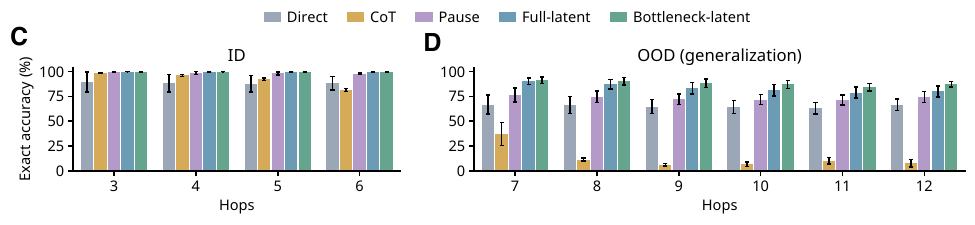}
\caption{\textbf{\pqt task and reasoning performance.}
\textbf{A} shows an example ($H=8$) with the correct path highlighted in green and distractor edges in gray.
\textbf{B} shows the five model variants.
\textbf{C} and \textbf{D} show free-generation accuracy on ID and OOD problems, respectively (error bars are SEM).}
\label{fig:task-model-perf}
\end{figure}

We train five model variants on the same \pqt dataset with the same tokenizer to examine how different forms of thinking address the symbolic reasoning problem.
All variants use the same GPTNeoX backbone (number of layers=4, hidden size=256, dimensionality of FFN=768).
Unlike \citet{haoTrainingLargeLanguage2025}, all models are trained from random initialization, ensuring
their knowledge of the task comes completely through training, rather than possibly inherited from pretraining.
The variants differ in their intermediate computation and answer readout (Fig.~\ref{fig:task-model-perf}B).
For all models, we denote the residual state at token position $i$ after block $\ell$ by $h_i^{(\ell)}$, and denote $\ell=0$ as the input to the first block.
During prompt encoding, the token embedding $E_\theta$ supplies $h_i^{(0)}=E_\theta(x_i)$.
After the final block, a final LayerNorm $N_\theta$ and an output projection map $h_i^{(L)}$ to next-token logits.

After the input $x$, the \drt variant generates the answer directly.
The \cotv variant first generates a proof and then the answer.
Its training is supervised by both the shortest proofs (a sequence of premises forming the shortest path from $r$ to $c^*$) and final answers.
In the \pstk variant, before answer decoding, the model ``thinks'' by inserting K=6 identical learnable embeddings, $z_t=E_\theta(\texttt{<PAUSE>})$.
The two latent variants, \flt and \btnlt, in their ``thinking'' process, instead, feed the normalized output of one step directly into the next, instead of decoding it into a token.
This process can be described with
\begin{equation}
        z_1 = N_\theta\!\left(h_n^{(L)}\right), z_{t+1} = N_\theta\!\left(h_{n+t}^{(L)}\right),
        \qquad 1\leq t<K, \label{eq:latent-feedback}
\end{equation}
in which each vector enters the first block as $h_{n+t}^{(0)}=z_t$, where $n=|x|$.

These two latent variants differ in the readout phase.
The \flt model processes $z_K$ with the prompt and recurrent cache retained, keeping both available during answer generation.
In this sense, it is similar to the Coconut model (\citet{haoTrainingLargeLanguage2025}), but without human guidance involved.
However, \flt can still access the $g$ and $q$ during answer decoding, which may allow the model to exploit statistical patterns to solve the problem.
To isolate the effect of direct access, we build the \btnlt model, which cannot access the recurrent key/value cache and has to re-encode the $z_{1:K}$ as the sole prefix for answer generation.
This allows us to examine what is happening inside the latent state trajectories and to decompose the reasoning with memorization.

\section{Results}
\subsection{ID and OOD dataset performance}
All models are trained on the same set of problems with $H\in\{3,\ldots,6\}$, using the same budget and optimization settings (see ~\ref{training_details}).
On the ID validation set, \btnlt and \flt achieve
nearly perfect accuracy, and \pstk is closely behind.
\cotv also achieves high accuracy overall, although its performance
declines as the $H$ increases.
In contrast, without additional computational slots,
\drt is worse than others and is more sensitive to the seeds.

We then test their generalization capability with OOD dataset ($H\in\{7,\ldots,12\}$).
Without further training, the variants start to show divergent behaviors.
\flt and \btnlt retain the highest accuracy, with \btnlt slightly ahead of \flt, and both clearly outperforming \pstk and \drt.
Unexpectedly, \cotv performs worst despite its strong ID performance.
Thus, models that perform similarly on the training range can generalize very differently beyond it (Fig.~\ref{fig:task-model-perf}C,D).

\subsection{How do different models solve the \pqt task?}
\subsubsection{Alignment with forward propagation}
The divergent OOD performance suggests that these variants may reach the same answer through different computations.
Two natural heuristic strategies are forward propagation from the query root and backward tracing from the candidate answers.
We first test whether their representations track forward propagation.
Using representational similarity analysis (RSA), we compare pairwise dissimilarities between model representations and algorithmic states, without assuming a one-to-one correspondence between model steps and algorithmic updates (Appendix~\ref{app:rsa}).

\begin{figure}[htbp]
\centering
\includegraphics[width=\linewidth]{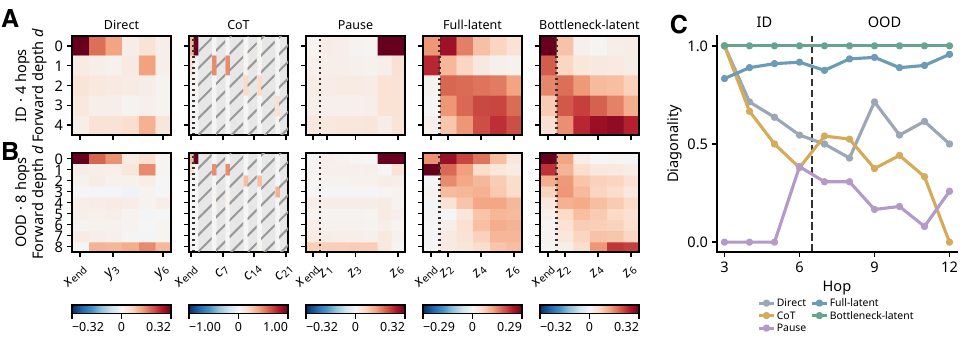}
\caption{\textbf{Representational alignment with forward graph propagation.}
\textbf{A} and \textbf{B} show RSA heatmaps for five variants on ID (4-hop) and OOD (8-hop) problems, respectively, using the same examples across variants.
Each entry shows the Spearman correlation between pairwise model-representation dissimilarities and pairwise graph-frontier dissimilarities at depth $d$.
For $G=(V,E)$ with query root $r$, the propagation frontiers are defined by
$F_0=\{r\}$ and
$F_{d+1}=\{v\in V:\exists u\in F_d,\ (u,v)\in E\}$.
Gray hatched cells denote undefined correlations.
Color scales are shared across rows within each variant.
\textbf{C} shows diagonality across 3--12-hop problems, with colors indicating variants.
The dashed line separates ID (3--6 hops) from OOD (7--12 hops).}
\label{fig:forward-rsa}
\end{figure}

\drt and \pstk show nearly no sequential alignment with forward propagation (Fig.~\ref{fig:forward-rsa}A,B).
Nevertheless, \cotv shows some alignment, consistent with its supervision on step-by-step proofs, despite very poor OOD performance.
The clearest diagonal-like patterns appear in \flt and \btnlt, suggesting that they learn a forward-search-like procedure without intermediate supervision.
We quantify this progression intuition with \emph{diagonality}, which measures whether alignment shifts toward later model positions as algorithm depth increases (Appendix~\ref{diagonality_def}).
A score approaching one indicates a consistent progression, without requiring one model step per graph hop.
\btnlt has the highest and most consistent diagonality across depths, followed by \flt, with the clearest separation from the other variants in OOD problems (Fig.~\ref{fig:forward-rsa}C).
The corresponding analysis of parallel backward tracing shows no clear, consistent sequential alignment across task depths (Appendix~\ref{app:rsa}, Fig.~\ref{fig:backward-diagonality}).

\subsubsection{Local graph shortcuts drive predictions in non-latent models}\label{shortcuts_analysis}
Latent models represent search-related intermediate variables, but no comparable alignment is found in \drt, \cotv, or \pstk. However, these three variants remained highly accurate within the training distribution, which raises the question of what supports their answers. One possibility is that they exploit shortcuts from local graph features. Such shortcuts are actually available since the graph-generation procedure inherited from ProsQA induces systematic degree asymmetries: correct candidates tend to have lower in-degree than incorrect ones, and the immediate successors of the query root that lead to the correct candidate tend to have lower in-degree and higher out-degree than the alternatives. These correlations provide local predictive cues, so we test whether model variants rely on them by manipulating local graph features.

\begin{figure}[htbp]
\centering
\includegraphics[width=\linewidth]{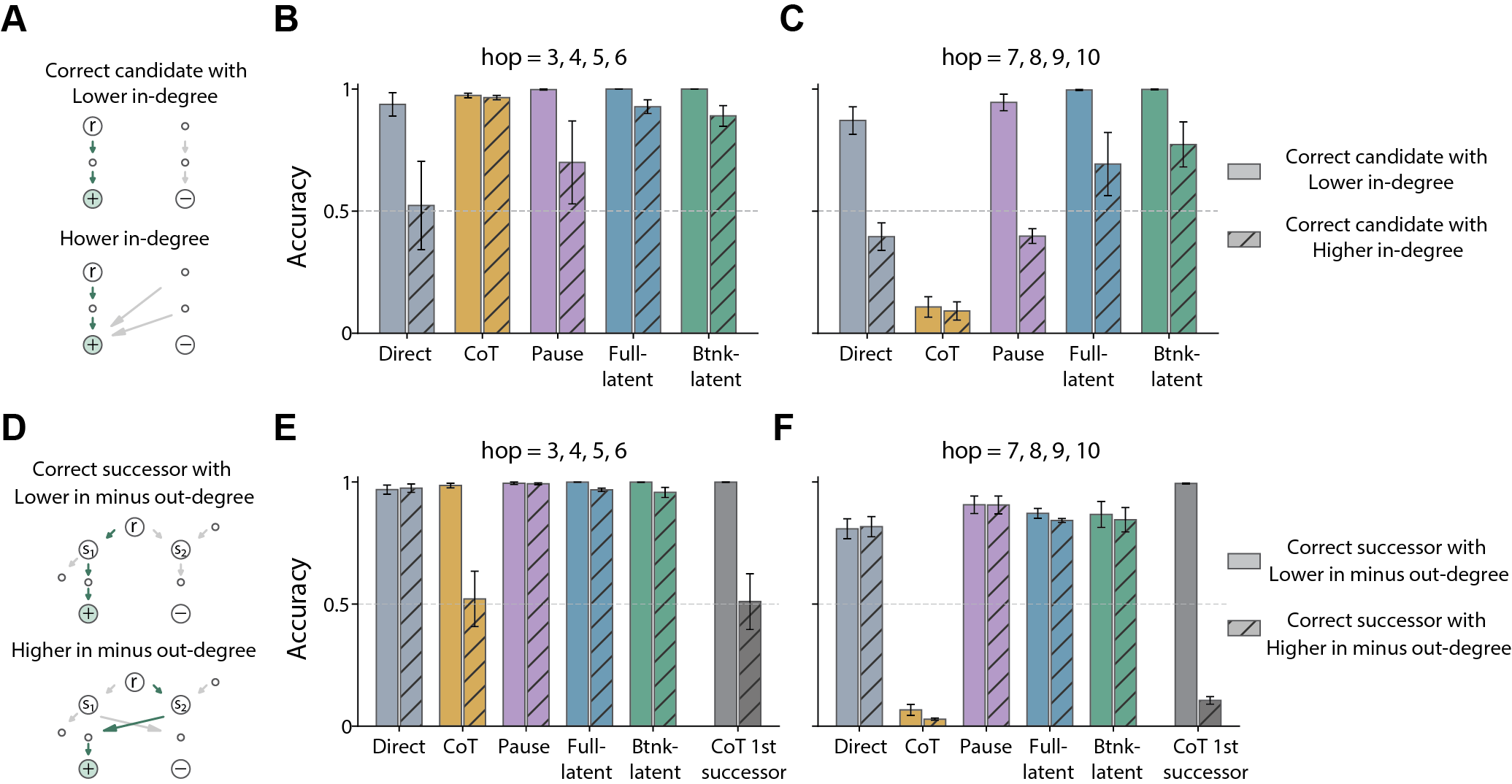}
\caption{\textbf{Effects of candidate and successor degree on model predictions.}
\textbf{A} shows matched pairs that reverse the candidates’ relative in-degree while preserving the proof path and correct answer. \textbf{B} and \textbf{C} show the final-answer accuracy on problems with short and long reasoning depths. Solid and hatched bars indicate that the correct candidate has lower and higher in-degree, respectively. \textbf{D} shows matched pairs that switch which immediate successor of the query root leads to the correct candidate while preserving all node degrees and the correct final answer. \textbf{E} and \textbf{F} show final-answer accuracy for all five models and first-successor accuracy for CoT on problems with short and long reasoning depths. Solid and hatched bars indicate that the correct successor has a lower or higher in-degree minus out-degree, respectively. In schematics, (r) denotes the query root, (+) and (-) denote the correct and incorrect candidates, and green arrows indicate the proof path.}
\label{fig:degree}
\end{figure}

We first ask whether the models use the in-degree of candidates when generating answers. To test this, we construct a matched-pair dataset, in which we select one or two edges not on the proof path and redirect their destinations to the correct candidate or to the incorrect candidate (Fig.~\ref{fig:degree}A). This matched pair differs only in the in-degrees of two candidates, keeping all others the same. Note that this matched-pair dataset is constructed to isolate individual graph features and differs in structure from the OOD evaluation set in Section 3.1. Evaluating the five variants on this dataset shows that candidate in-degree strongly affects accuracy in \drt and \pstk across the problems with different reasoning depths, but has much smaller effects on the latent models and minimally affects \cotv (Fig.~\ref{fig:degree}B, C).

Next, we ask whether the models use local degree cues at the query root’s immediate successors. Unlike other variants, \cotv explicitly identifies a successor in its first generated statement before producing the final answer, making it more vulnerable to the degree at the query root’s successors. To test this, we construct a second matched-pair dataset in which the query root has two immediate successors, with one leading to the correct candidate. Within each pair, we change the in/out degrees of successors by reassigning the source endpoints of non-proof outgoing edges from one successor to the other and redirecting the destination endpoints of non-proof incoming edges between them (Fig.~\ref{fig:degree}D). These manipulations strongly affect both the first-step successor choice and the final answer in \cotv, which tends to prefer the successor with lower in-degree minus out-degree, while having little effect on all other variants (Fig.~\ref{fig:degree}E, F).

Together, these results indicate that \drt and \pstk rely substantially on candidate in-degree when predicting the answer, whereas \cotv uses the degree of the query root's successors both when selecting which successor to follow and when predicting the answer. In contrast, the latent models are less affected by either type of local graph structure.

\subsection{A recurrent search algorithm in latent reasoning models}

\subsubsection{Swapped chains as a probe of a ``soft'' for-loop}
Analyses in Fig.~\ref{fig:forward-rsa} suggest that, unlike other variants, latent reasoning models, especially \btnlt, may implement some recurrent algorithms that support generalization to longer hop problems.
To further show if this causally holds, we swap the graph connectivity at each depth and use the change as a probe.

\begin{figure}[htbp]
\centering
\includegraphics[width=1.0\linewidth]{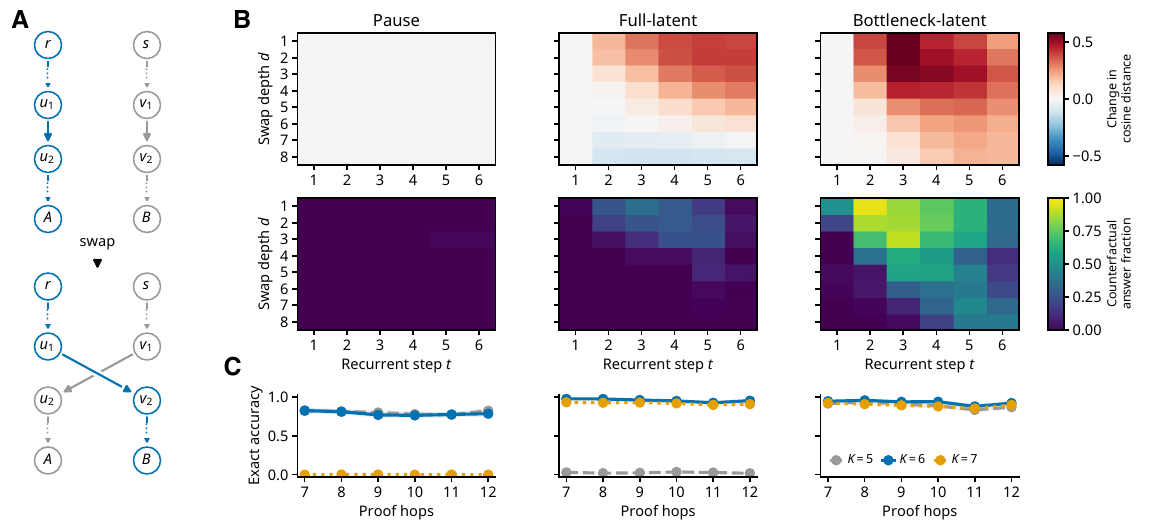}%
\caption{\textbf{Controlled interventions probe the recurrent computation in latent reasoning models.}
\textbf{A} shows how swapping premises shifts the query root $r$'s correct candidate from \texttt{A} to \texttt{B}.
In \textbf{B}, the top row shows the normalized cosine distance between paired latent states for 8-hop problems with connectivity swaps at depth $d$, and the bottom row shows the fraction of pairs for which transplanting the latent state at step $t$ redirects the answer to $y'$.
\textbf{C} shows OOD accuracy with one fewer ($K=5$), the trained number ($K=6$), or one additional ($K=7$) thinking position.
Shaded region indicates 95\% CI.}
\label{fig:causal-intervention}
\end{figure}

For each sample $x=g\Vert q$, we build a matched one $x'=g'\Vert q$ by swapping the destinations of two edges at depth $d$: one on the solution chain and one on a matched distractor chain.
This preserves the query root, node labels, premise order, and node degrees, but switches the reachable candidates, flipping the answer from $y$ to $y'$ (Fig.~\ref{fig:causal-intervention}A).
Next, we run both inputs to obtain the latent trajectories $z_{1:K}(x)$ and $z_{1:K}(x')$.
At step $t$, we replace $z_t(x)$ with $z_t(x')$, keep the original prompt cache and preceding computation unchanged, and recompute the remaining latent states before generating the answer.
We apply this probe to \btnlt and \flt, with \pstk as a control.
We measure the state-level effect of the connectivity change as the cosine distance between $z_t(x)$ and $z_t(x')$, normalized by subtracting the distance at $t=1$ (Fig.~\ref{fig:causal-intervention}B top).
The causal influence is the fraction of pairs for which transplantation changes the answer from $y$ to $y'$ (Fig.~\ref{fig:causal-intervention}B bottom).

If the recurrent computation propagates reachability step by step, shallow perturbations should become effective earlier, and deep ones should influence the output only at later steps.
This upper triangular pattern clearly emerges in \btnlt (Fig.~\ref{fig:causal-intervention}B,C), while it is less clear in \flt and absent in \pstk.
These results suggest that, among the three variants, only \btnlt strongly adopts a recurrent algorithm.

Nevertheless, such computation might be ``soft'' rather than a strict ``hard'' for-loop, since a single latent step handles more than one exact hop.
This soft-iteration hypothesis makes a further prediction: small perturbations to $K$ should barely affect \btnlt's accuracy.
Indeed, only \btnlt retains its performance under $K=5,6,7$, whereas \flt is sensitive to $K=5$ and \pstk to $K=7$ (Fig.~\ref{fig:causal-intervention}C).
Together, these results support that \btnlt may implement a ``soft'' forward search through its recurrent circuit.

\subsubsection{Localize the sparse circuit inside the bottleneck latent model}

\begin{figure}[htbp]
\centering
\includegraphics[width=\linewidth]{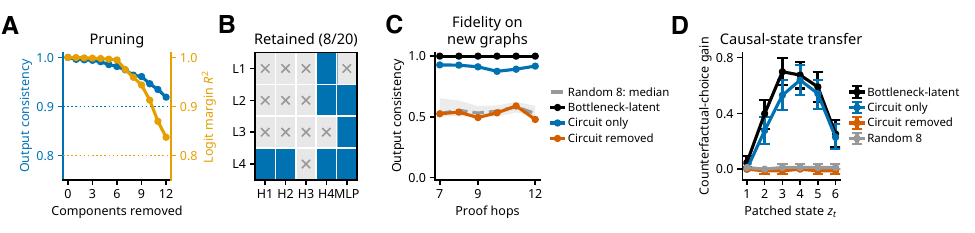}
\caption{\textbf{Localization of the recurrent circuit in \btnlt.}
Replacement based pruning (\textbf{A}) retains eight of twenty recurrent components (\textbf{B}).
The selected circuit largely preserves candidate choices (\textbf{C}) and causal state-transfer effects (\textbf{D}), whereas removing it or retaining random size-matched components does not.
Error bars indicate pointwise 95\% bootstrap confidence intervals over base graphs; the gray band shows the 10th--90th percentiles across twenty random circuits.}
\label{fig:bottleneck-circuit}
\end{figure}

Motivated by the RSA and intervention analyses above, we examine which components support the recurrent computation in \btnlt and how this circuit enables OOD generalization (Fig.~\ref{fig:bottleneck-circuit}) by circuit pruning to the recurrent updates from $z_1$ to $z_K$.
We check all 20 physical components in \btnlt, including $4\times$ attention heads and $1\times$ MLP per layer in GPTNeoX.
A component is removed if the remaining circuit retains at least 90\% output consistency with the full model and an $R^2>0.8$ for the candidate logit margin.
This procedure (see~\ref{app:circuit-pruning}) greedily continues until no further component can be removed (Fig.~\ref{fig:bottleneck-circuit}A), leading to an 8-component sparse circuit (Fig.~\ref{fig:bottleneck-circuit}B) that preserves 91.9\% output consistency and an $R^2$ of $0.837$.

We next run the pruned circuit on other 7 to 12-hop samples that are not involved in pruning.
The circuit retains 90.9\% output consistency with the full model, above the 52.7\% when these eight components are removed and 55.8\% when a random size-matched subset is retained instead (Fig.~\ref{fig:bottleneck-circuit}C).

To test whether the selected circuit preserves the causal recurrent mechanism in \btnlt, we repeat the latent-state transplantation.
On separate 8-hop pairs with a connectivity swap at depth 4, we measure the increase in counterfactual-answer choices relative to each condition's own baseline.
The selected circuit retains a similar step-dependent transfer profile of the full model, with a peak increase of 63.9\% versus 69.9\%.
In contrast, this effect is largely absent when the circuit is replaced by a random size-matched subset (Fig.~\ref{fig:bottleneck-circuit}D).
In addition, the results remain stable across 7-12 hops (Fig.~\ref{fig:moving-perturbation-pruned}).
Together, our selected circuit preserves not only the
model's output, but also the causal state-transfer mechanism identified above.

\subsubsection{A recurrent search algorithm inside the sparse circuit}
With the pruned circuit narrowing the recurrent computation to a 8 components, we next examine the specific role of each during recurrent computation.
Among them, two components are particularly interesting. Attention head 1 in layer 4 (L4H1) separates how premise sources and destinations are transmitted. The corresponding MLP (L4MLP) helps propagate the retrieved information into subsequent recurrent states.

\begin{figure}[htbp]
\centering
\includegraphics[width=\linewidth]{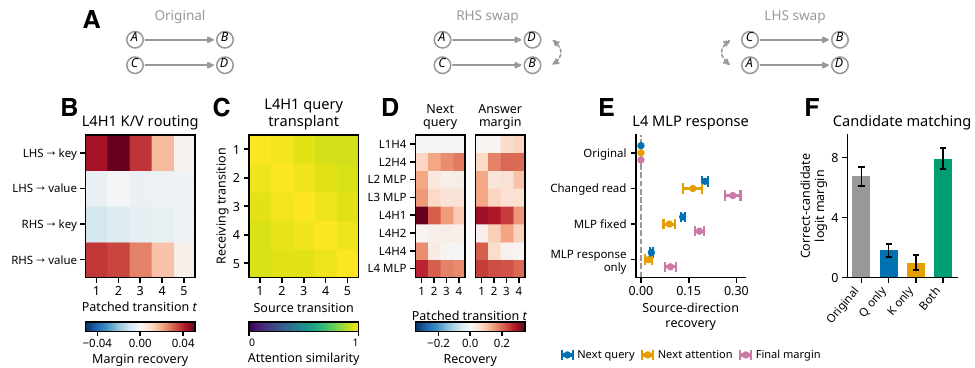}
\caption{\textbf{Functional analysis of the recurrent circuit in \btnlt.}
\textbf{A} illustrates the LHS and RHS swaps.
Interventions within the pruned circuit distinguish L4H1's key/value routing (\textbf{B}), query-dependent selection of reading depth (\textbf{C}), and component contributions to the next query and final answer (\textbf{D}).
Fixing or transplanting the L4 MLP response (\textbf{E}) separates its contribution from the residual pathway.
\textbf{F} tests candidate matching through Q/K interventions in L4H1/H2/H4.
Error bars indicate pointwise 95\% bootstrap confidence intervals.}
\label{fig:bottleneck-circuit-roles}
\end{figure}

For a premise \texttt{A is B.}, A and B are referred to as the left-hand side (LHS) and the right-hand side (RHS), respectively.
Using the same constructions above, we create the same candidate-switching interventions by either swapping the LHS or the RHS at the same premise location (Fig.~\ref{fig:bottleneck-circuit-roles}A).
By replacing the actual cache with the swapped one, this matched-pair swap isolates whether the influence propagates through the RHS K-cache or the V-cache.
We first characterize how attention reads graph premises.
We find that, in L4H1, LHS swaps affect the answer mainly through keys, whereas RHS swaps affect mainly through values (Fig.~\ref{fig:bottleneck-circuit-roles}B).
To check if this division persists across different steps, we measure the similarity between the attention distribution before and after transplant using Jensen-Shannon distance, and find that these distributions are highly consistent ($\approx 1$, Fig.~\ref{fig:bottleneck-circuit-roles}C).
These results suggest an 'address--content'  organization in L4H1: keys determine which nodes are linked, while values supply the destination node information.

Then, we examine how MLP layers contribute to the computation.
Instead of changing the structure of $G$, we swap the query root node $r$ in matched chains used by Fig.~\ref{fig:causal-intervention}, and measure how each component shifts the next query and the final answer in each transition (Fig.~\ref{fig:bottleneck-circuit-roles}D).
The L4MLP along with L4H1 show persistent influence on both, suggesting that they work together and change the query direction to further influence the next recurrent step.
To isolate the contribution of L4MLP, we select the recurrent update $z_2 \to z_3$, use the swapped L4H1 and measure how L4MLP influences the downstream targets from next query to final answers  under different interventions.
With L4H1's output changed, we find recomputing L4MLP shifts all downstream targets to the swapped direction, compared with a frozen L4MLP (Fig.~\ref{fig:bottleneck-circuit-roles}E).
However, this effect is not additive (single L4MLP change barely shifts the direction) and relies on the residual stream (shifts exist even the MLP is fixed, though the magnitude is much lower).
This shows that, L4MLP, they works by the whole residual stream, act as a ``filter'', to select information for the next round's operation.

We next ask how the evolving $z_t$ becomes evidence for answer candidate $c^*$, given an $r$.
Candidate positions in $x$ carry graph conditioned representations that recurrent attention can read (during reasoning phase).
Using the sparse pruned circuit above, we keep the original question intact and test two types of changes with L4H1/H2/H4 during $z_t \to z_{t+1}$.
In the first change case, we replace the query with the Q from same step's update of a separate run starting from the other chain's root.
For example, if the original graph contains $r\leadsto A$ and $s\leadsto B$, we replace $Q_{r,t}$ with $Q_{s,t}$, while the original question still asks about $r$.
In the second condition, we use candidate keys obtained from a graph with exchanged candidate endpoints: $r\leadsto B$ and $s\leadsto A$, while keeping the question and its candidate positions unchanged.
Either change alone reduce the $c^*$ logit margin, while applying both changes together can restore it.
This suggests the candidate evidence depends on a match between the current recurrent state and the candidates' graph context.
Among the retained L4 Hs, H1 shows the strongest recurrent Q/K matching effects, whereas H2 shows the largest accuracy loss under candidate value exchange and the greatest logit margin recovery at final readout phase.

Together, these results reveal how a recurrent search algorithm is implemented in the \btnlt: the $z_t$ maintains currently reachable node in $G$; attention head L4H1 works as a soft tracer and uses premise LHSs to retrieve their RHSs to expand the reachable set;
and L4 MLP, together with the residual pathway, then incorporates the retrived information back to $z_{t+1}$, guiding the next round of seaching.
The candidate reading pathways in L4H1/H2/H4 also connect this evolving state to answer evidence.
The Q/K matching controls the candidate information written into the latent $z_t$, from which the final answer is decoded.
These processes can operate in parallel, which allows the \btnlt to build a faster reachability search than strict step-by-step traversal as we have seen in 7 to 12 hop problems.

\section{Conclusion}
We ask whether models under different forms of thinking develop mechanistically distinct solutions, or converge on the same solutions through different ways. We train five GPT-like variants with the same backbone on an extended ProsQA task, and compare the mechanisms they induce.

Similar in-distribution performance hides the mechanistic divergence among the models. \drt, \cotv, and \pstk models solve in-distribution problems well, but rely on shortcuts related to local graph structure and generalize poorly to out-of-distribution problems. Notably, \cotv, which is explicitly supervised with step-by-step proofs, fails to generalize, suggesting that training on reasoning traces does not guarantee that a model will reason in the same way. In contrast, the latent models (\flt and \btnlt), despite receiving no intermediate reasoning trace or reward signal, develop a recurrent circuit that implements ``soft'' forward search algorithm, expanding the reachable set across recurrent steps and generalizing to problems with longer reasoning depths.
Digging deeper into the circuit in the bottleneck model, we find that an attention head retrieves graph relations through an address–content organization, while an MLP, together with the residual stream, integrates the retrieved information into the next state and a multi-head reading mechanism performs the candidate matching.

Together, these results show that different thinking interfaces lead to distinct underlying mechanisms, even at similar performance. The latent-reasoning models that allow information to flow fully across steps develop genuine reasoning computation that matches the structure of the task.

\section{Discussion}
We want to emphasize the importance of testing model behavior on OOD problems before turning to internal mechanisms. As the Stroop's color–word interference task reveals how humans process language and visual information, carefully designed OOD problems can also reveal how a model solves a task and guide where further mechanistic analysis should go \citep{friedman2024interpretability}.

Building on this behavioral comparison, our design also isolates the effect of the thinking interface itself. Most mechanistic studies analyze a single model or a single form of reasoning in isolation. Instead, we train five variants that share the same backbone, dataset, and budget and differ only in their thinking interface, separating the mechanistic differences attributed to the interface from others. One family that is commonly used but not included here is the looped transformer, which applies the same weight-tied block for multiple iterations \citep{dehghaniUniversalTransformers2019,giannouLoopedTransformers2023}. This explicit recurrence formalizes the iterative computation that we discover in the latent models, but whether the latent models generate the same forward search is still left to future work.

Finally, our conclusions come from small models on a controlled symbolic reasoning task, which simplifies analysis and lets us narrow the computation down to an interpretable circuit with confidence. However, the use of small models and a synthetic task limits how far the current conclusions can extend. Thus, whether our results and conclusions hold for larger pretrained LLMs and more naturalistic problems remains an open question for future work.

\bibliography{iclr2027_conference}
\bibliographystyle{iclr2027_conference}

\appendix
\counterwithin{figure}{section}
\section{Appendix}

\subsection{Dataset Generation algorithm}
\label{dataset_generation_algo}

We construct \pqt from the graph-reachability task in ProsQA \citep{haoTrainingLargeLanguage2025}.
In ProsQA, each sample consists of a series of premises describe a DAG and a question asks which of the two candaites is reachable from the specified root node.
In \pqt, we preserved this logic, while we changed the how each sample is constructed.
Labels are assigned after graph construction, premises are shuffled, and the correct candidate appears equally often on either side of the question, so that no obvious superficial statistics are related to the answer.

For the extended 400k 3--6 hop training set, we sample new graphs with an empirical quota that keeps the empirical joint distribution of graph size, proof length, queried root, and binned number of shortest paths similar to ProsQA.
For ID validation and test, we retain the released graphs and queries, but reassign node labels and randomize premise and candidate order.

For OOD 7--12 hop dataset for evaluation, we use algorithm ~\ref{alg:controlled-long-hop} to generate it:
\begin{algorithm}[H]
\caption{Controlled long-hop graph generation}
\label{alg:controlled-long-hop}
\begin{algorithmic}[1]
\Require Proof length $H\in\{7,\ldots,12\}$; balanced candidate-side assignment $b$
\State Construct two vertex-disjoint directed paths $P_1,P_2$, each of length $H$
\For{$j\in\{1,2\}$}
    \State Add $13-H$ off-path nodes $U_j$
    \State Attach the first node of $U_j$ to the root of $P_j$
    \State Attach each remaining node of $U_j$ to an internal node of $P_j$
           or an earlier node of $U_j$
\EndFor
\State Sample additional edges into off-path nodes, without duplicates or cycles,
       until $|E|=38$
\State Choose one of $P_1,P_2$ as the queried component
\State Set $r$ to its root, $c^+$ to its endpoint, and $c^-$ to the endpoint
       of the other path
\State Randomly assign node labels and permute the premise order
\State Place $c^+$ on side $b$ of the binary question
\end{algorithmic}
\end{algorithm}

\subsection{Training details}\label{training_details}
All five variants are trained from random initialization on the same 400k \pqt problems with proof lengths of 3–6 hops, separately on two Nvidia 5090 and 4090 machines.
Each model uses a four-layer GPTNeoX backbone with hidden size 256, four attention heads, feed-forward dimension 768, untied input and output embeddings, and attention and hidden dropout of 0.1.
We use AdamW with a constant learning rate of $4\times10^{-4}$, $(\beta_1,\beta_2)=(0.95,0.999)$, weight decay of $10^{-4}$ and a batch size of 256.
Training used BF16 without learning-rate warmup or decay.
The objective averages next-token cross-entropy over prompt and answer tokens with equal per-token weights;
Text CoT additionally supervised the shortest-proof tokens. Pause and latent variants used $K=6$ thinking positions.
Latent models are optimized end-to-end through the full recurrence without discrete intermediate targets, curriculum training or RL.

\subsection{Observed correlation between models' internal computation and the algorithm intermediate variables}
\label{app:rsa}

We compare model representations with propagated frontiers under forward search and successive ancestor sets of the candidates under backward tracing.
For a graph $G=(V,E)$ with query root $r$, the forward frontier is initialized at the root and updated by following outgoing edges:
\[
F_0=\{r\}, \qquad
F_{d+1}=\{v\in V:\exists u\in F_d,\ (u,v)\in E\}.
\]
Each update expands all nodes in the current frontier, allowing multiple branches to be followed in parallel.
Thus, $F_d$ contains nodes reachable from $r$ by a directed path of exactly $d$ edges.
It differs from the cumulative reachable set $\bigcup_{j=0}^{d}F_j$, which retains nodes reached at earlier depths.
A node can appear in multiple frontiers if paths of different lengths lead to it.
For example, edges $r\to a$, $r\to b$, and $a\to b$ give $F_1=\{a,b\}$ and $F_2=\{b\}$.

For backward tracing, we start from both candidates simultaneously and propagate their joint frontier along incoming edges:
\[
U_0=\{c_0,c_1\}, \qquad
U_{d+1}=\{u\in V:\exists v\in U_d,\ (u,v)\in E\}.
\]
Here, $U_d$ is the union of both candidates' frontiers after exactly $d$ reverse steps.
For RSA, forward and backward frontiers are encoded as binary vectors over node labels, indicating membership in the set.
The backward representation is invariant to the order of candidates in the query.

\begin{figure}[H]
\centering
\includegraphics[width=0.6\linewidth]{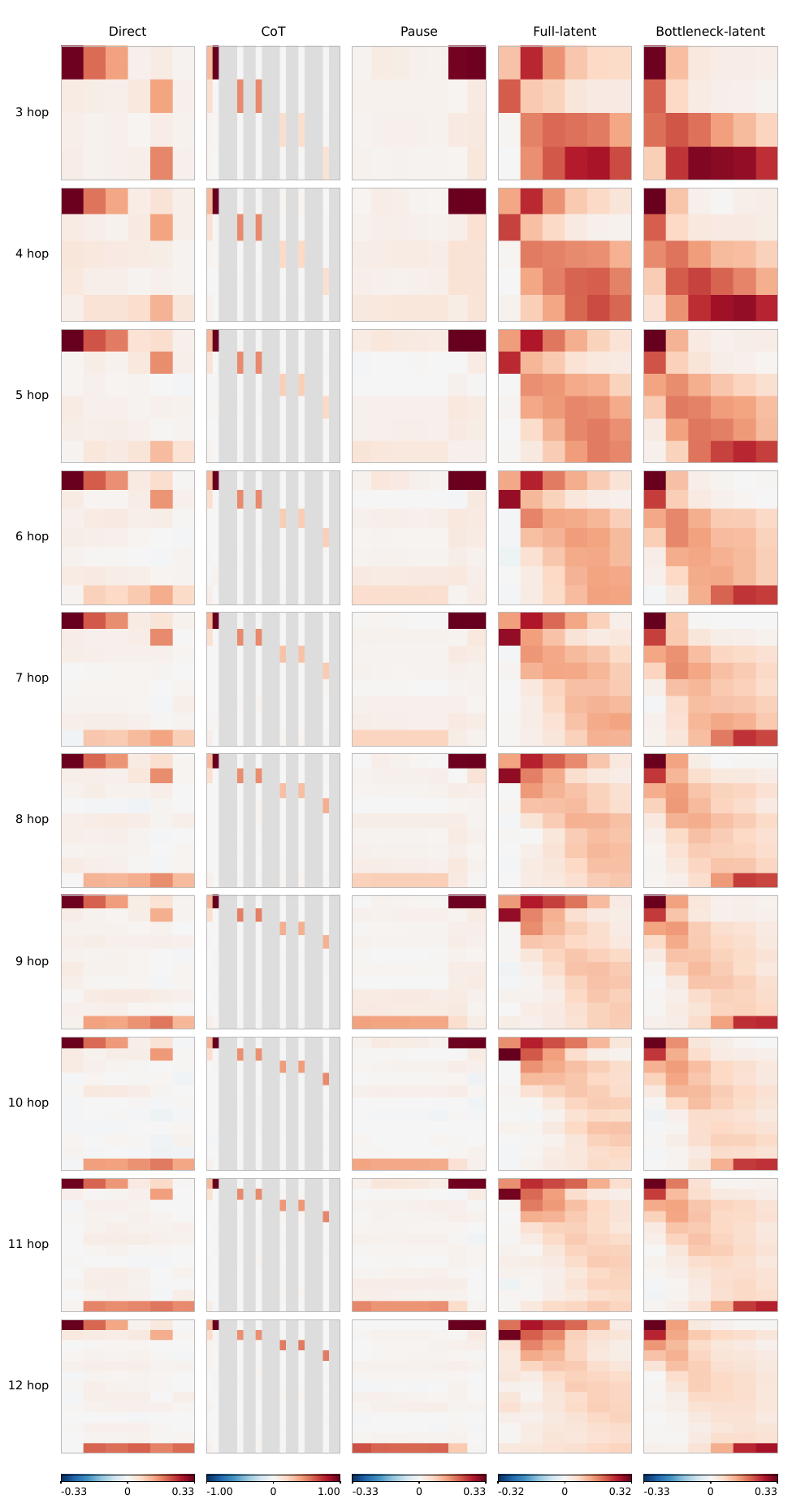}
\caption{\textbf{Representational alignment with forward graph propagation across task depths.}
RSA heatmaps for five model variants on 3 to 12-hop problems (rows), across ID (3--6 hops) and OOD (7--12 hops) conditions.
Entries represent Spearman correlations between model-representation and frontier dissimilarities, as shown in Fig.~\ref{fig:forward-rsa}.}
\label{fig:forward-rsa-all-hops}
\end{figure}

The analysis covers the final prompt position and the thinking phase, where applicable, on both ID and OOD problems.
For latent models, we use the continuous states $z_t$, with $z_1$ defined as the normalized last-layer state at the final prompt position.
For \cotv, we use the first 21 generated proof tokens, matching the shortest generated proof among the samples.
For \pstk, we analyze the normalized last-layer residual states $N_\theta(h_{n+t}^{(L)})$, since its input vectors $z_t$ are identical across steps.
\drt serves as a control without additional thinking steps, using normalized last-layer states at the final prompt position and during answer readout.
Figure~\ref{fig:forward-rsa-all-hops} shows forward-propagation alignment across all evaluated depths.

\subsection{Diagonality}\label{diagonality_def}
We develop \emph{Diagonality} to quantify whether stronger alignment shifts toward later computation
positions as algorithm depth increases.
Let $R_d(t)$ denote the RSA value at depth $d$ and position $t$.
First, we convert each row to normalized ranks,
$r_d(t)=(\operatorname{rank}(R_d(t))-1)/(n_d-1)$,
where $n_d$ counts finite entries.
Undefined entries and rows with fewer than two finite entries are set to zero.

We the compute the best fixed position, independently selected row maxima,
and the best nondecreasing path:
\begin{equation}
S_{\mathrm{static}}=\max_t\sum_d r_d(t), \qquad
S_{\mathrm{top}}=\sum_d\max_t r_d(t), \qquad
S_{\mathrm{fwd}}=
\max_{t_0\leq\cdots\leq t_D}\sum_{d=0}^{D}r_d(t_d).
\end{equation}
We define \emph{diagonality} as
\begin{equation}
\mathrm{Diagonality}
=
\frac{S_{\mathrm{fwd}}-S_{\mathrm{static}}}
     {S_{\mathrm{top}}-S_{\mathrm{static}}}.
\end{equation}
The score lies in $[0,1]$: when the denominator is positive, one means
that a nondecreasing path reaches every row's maximum (perfect propagation), while zero means
that allowing forward propagation gives no advantage over a fixed position.
We assign zero when the denominator vanishes.

Note that \emph{Diagonality} measures the ordering of alignment, not its absolute
strength. It allows pauses and jumps between positions, without
requiring one model step per algorithm update.

\begin{figure}[htbp]
\centering
\includegraphics[width=0.6\linewidth]{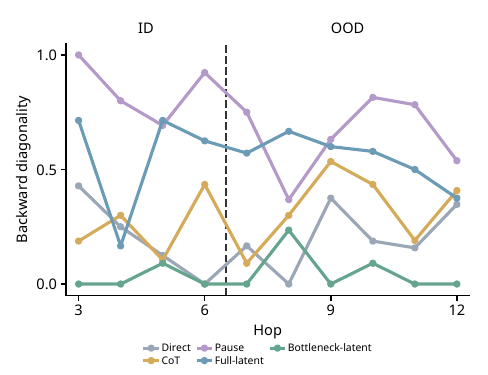}
\caption{\textbf{Diagonality of alignment with parallel backward search.}
The same diagonality measure is applied to RSA against the joint backward frontier for all five model variants.
The dashed line separates ID from OOD problems.}
\label{fig:backward-diagonality}
\end{figure}

Across numer of hops, no variant shows a clear, consistent pattern of alignment with backward search (Fig.~\ref{fig:backward-diagonality}).

\subsection{Effect of hop perturbation on pruned circuit in Bottleneck-latent model}
\begin{figure}[H]
\centering
\includegraphics[width=1.0\linewidth]{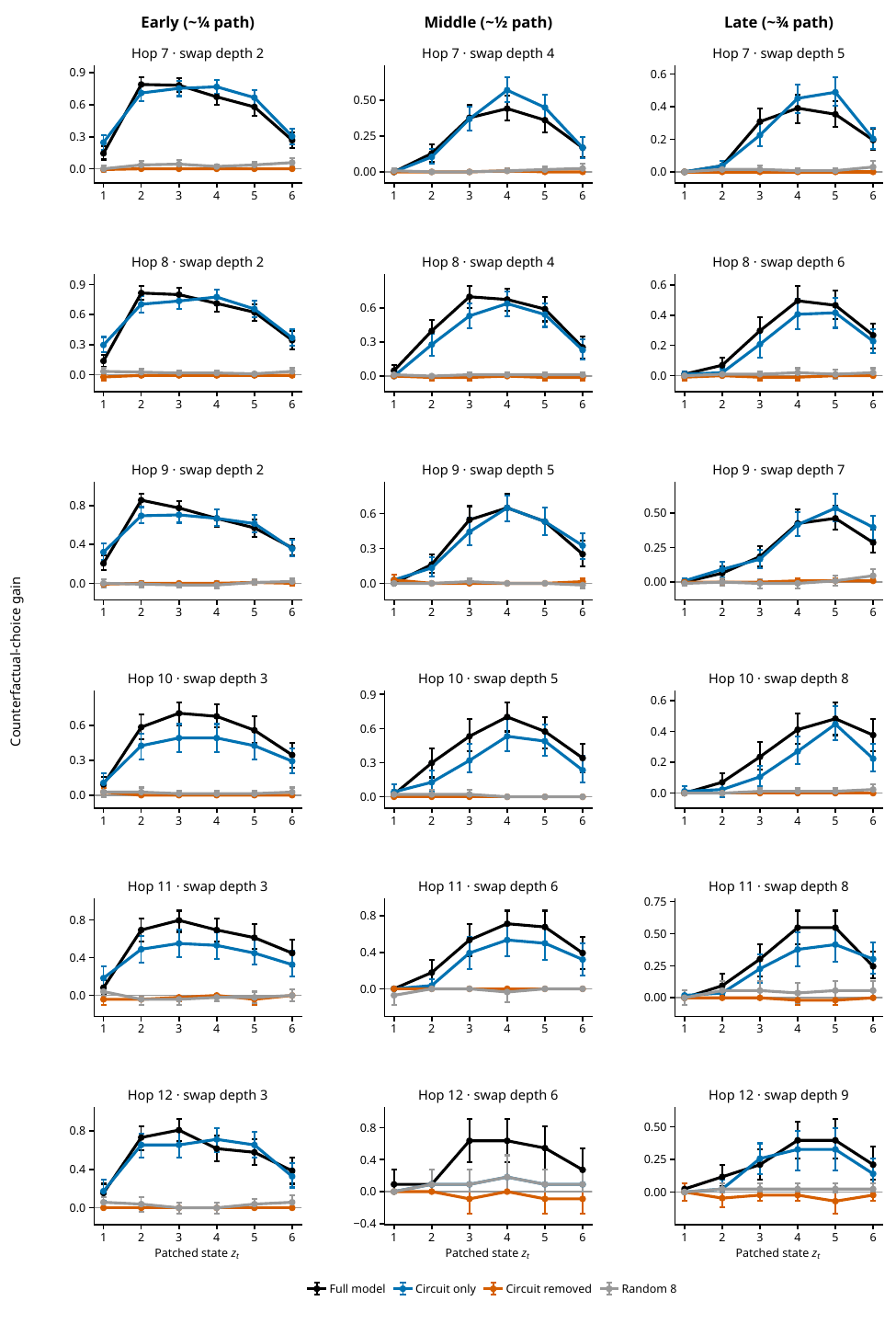}
\caption{\textbf{Effect of hop perturbation on pruned circuit in Bottleneck-latent model.}}
\label{fig:moving-perturbation-pruned}
\end{figure}

Figure~\ref{fig:moving-perturbation-pruned} shows the effect of hop perturbation on state transfer in the pruned circuit across 7--12-hop problems.

\subsection{Recurrent circuit pruning}
\label{app:circuit-pruning}

We apply greedy mean-replacement pruning (Algorithm ~\ref{alg:circuit-pruning}) to the 20 recurrent components of \btnlt (16 attention heads and four MLPs).
Each component is pruned across all five recurrent transitions from $z_1$ to $z_6$.
Replacement means are computed separately for each component and transition from both members of counterfactual pairs constructed from 512 independent 8-hop graphs.
These means remain fixed throughout pruning.
Input encoding and answer readout remain the same.

For a retained component set $S$, let $m_i(S)$ denote the candidate logit
margin after mean-replacing all components outside $S$, and
let $m_i$ denote the full-model margin.
We measure candidate-choice consistency $C(S)$ and margin
fidelity $R^2(S)$ on the examples:
\[
C(S)=\frac{1}{N}\sum_i
\mathbf{1}\!\left[
\mathbf{1}[m_i(S)\geq0]=\mathbf{1}[m_i\geq0]
\right],
\qquad
R^2(S)=1-
\frac{\sum_i(m_i(S)-m_i)^2}
     {\sum_i(m_i-\bar m)^2}.
\]

\begin{algorithm}[htbp]
\caption{Greedy recurrent circuit pruning}
\label{alg:circuit-pruning}
\begin{algorithmic}[1]
\Require Full component set $\mathcal{C}$; fixed replacement means
\State $S \gets \mathcal{C}$
\While{$S\neq\varnothing$}
    \For{each $c\in S$}
        \State Evaluate $C_c=C(S\setminus\{c\})$ and
               $R_c^2=R^2(S\setminus\{c\})$
    \EndFor
    \State $\mathcal{F}\gets
        \{c\in S:C_c\geq0.90,\ R_c^2\geq0.80\}$
    \If{$\mathcal{F}=\varnothing$}
        \State \textbf{break}
    \EndIf
    \State $c^\star\gets\displaystyle\arg\min_{c\in\mathcal{F}}
        \frac{1}{2}\left(
        \frac{1-C_c}{0.10}+\frac{1-R_c^2}{0.20}\right)$
    \State $S\gets S\setminus\{c^\star\}$
\EndWhile
\State \Return $S$
\end{algorithmic}
\end{algorithm}

\subsection{Head contributions during recurrence and answer readout}
\label{app:candidate-heads-by-stage}

We compare L4's H1, H2, and H4 using the same \btnlt  recurrent circuit (Fig~\ref{fig:bottleneck-circuit}).

During recurrence, we apply the Q/K interventions in Fig.~\ref{fig:bottleneck-circuit-roles}F to one head at a time across all five updates. We measure the decrease in the  $correct - incorrect$  candidate logit margin.
We also test candidate values separately: we reverse the candidate order in a separate run and use its candidate position values in the original run, keeping the attention weights unchanged at each update. We measure the resulting accuracy loss in percentage points.

For final answer readout, we use graph pairs with opposite correct answers but the same task root, candidate positions, and answer prefix.
We replace one head's output at the answer prediction position with its output from the opposite-answer run.
The original latent trajectory stays fixed.
43 of 96 graph pairs for which both answers are initially predicted correctly are used for analysis.
Here, the margin is the right candidate's logit - the remaining candidate's logit.
Recovery measures the shift toward the opposite answer margin: 0\% means no change, and 100\% means reaching that margin.

\begin{figure}[htbp]
\centering
\includegraphics[width=\linewidth]{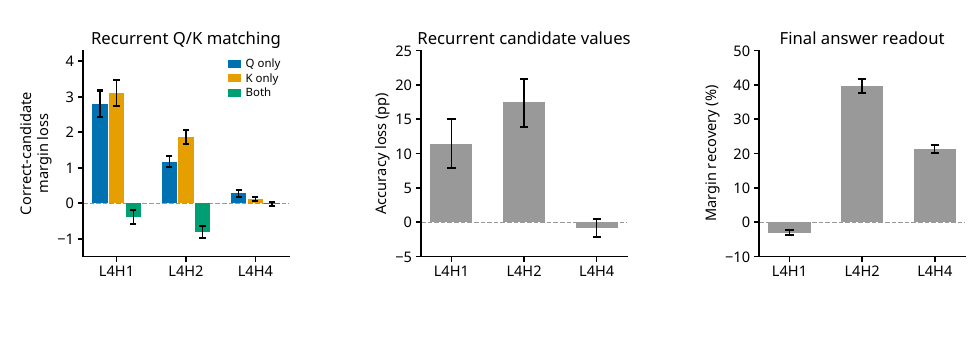}
\caption{\textbf{Head contributions during recurrence and answer readout.}
Individual interventions compare recurrent Q/K matching (left), candidate-value exchange (middle), and final answer readout (right) for L4H1, L4H2, and L4H4.
Error bars indicate pointwise 95\% bootstrap confidence intervals.}
\label{fig:candidate-heads-by-stage}
\end{figure}
\end{document}